\documentclass[10pt]{article} 
\usepackage[accepted]{tmlr}

\usepackage{microtype}
\usepackage{graphicx}
\usepackage{subfigure}
\usepackage{booktabs} 

\usepackage{hyperref}
\usepackage{url}

\usepackage{algorithm}
\usepackage{algpseudocode}
\usepackage{eqparbox}
\usepackage[normalem]{ulem}
\newdimen{\algindent}
\algnewcommand\LeftComment[2]{%
   \hspace{#1\algindent}$\triangleright$ \eqparbox{COMMENT}{#2} \hfill %
}

\makeatletter
\newcommand{\disablepackage}[2]{%
  \disable@package@load{#1}{#2}%
}
\makeatother
\disablepackage{algorithmic}{}

\usepackage{amsmath}
\usepackage{amssymb}
\usepackage{mathtools}
\usepackage{amsthm}

\usepackage[capitalize,noabbrev]{cleveref}

\usepackage{enumitem}
\usepackage{graphicx}

\graphicspath{{./}{graphics/}}

\usepackage{amsmath,amsfonts,bm}

\def\eqref#1{equation~\ref{#1}}

\def\1{\bm{1}}

\DeclareMathAlphabet{\mathsfit}{\encodingdefault}{\sfdefault}{m}{sl}
\SetMathAlphabet{\mathsfit}{bold}{\encodingdefault}{\sfdefault}{bx}{n}

\newcommand{\softmax}{\mathrm{softmax}}

\DeclareMathOperator*{\argmax}{arg\,max}

\DeclareMathOperator{\logsumexp}{logsumexp}
\DeclareMathOperator{\entropy}{entropy}
\DeclareMathOperator{\softmaxentropy}{sm\_entropy}

\definecolor{mybrown}{RGB}{165, 42, 42}

\title{AutoCLIP: Auto-tuning Zero-Shot Classifiers for Vision-Language Models}

\author{\name Jan Hendrik Metzen \email JanHendrik.Metzen@de.bosch.com \\
      \addr Bosch Center for Artificial Intelligence, Robert Bosch GmbH
      \AND
      \name Piyapat Saranrittichai \email Piyapat.Saranrittichai@de.bosch.com \\
      \addr Bosch Center for Artificial Intelligence, Robert Bosch GmbH
      \AND
      \name Chaithanya Kumar Mummadi \email 
      ChaithanyaKumar.Mummadi@de.bosch.com\\
      \addr Bosch Center for Artificial Intelligence, Robert Bosch LLC}

\def\month{08}  
\def\year{2024} 
\def\openreview{\url{https://openreview.net/forum?id=gVNyEVKjqf}} 

\begin{document}
\maketitle

\begin{abstract}

Classifiers built upon vision-language models such as CLIP have shown remarkable zero-shot performance across a broad range of image classification tasks. Prior work has studied different ways of automatically creating descriptor sets for every class based on prompt templates, ranging from manually engineered templates over templates obtained from a large language model to templates built from random words and characters. Up until now, deriving zero-shot classifiers from the respective encoded class descriptors has remained nearly unchanged, i.e., classify to the class that maximizes cosine similarity between its averaged encoded class descriptors and the image encoding. However, weighing all class descriptors equally can be suboptimal when certain descriptors match visual clues on a given image better than others. In this work, we propose \textsc{AutoCLIP}, a method for \emph{auto-tuning zero-shot classifiers}. \textsc{AutoCLIP} tunes per-image weights to each prompt template at inference time, based on statistics of class descriptor-image similarities. \textsc{AutoCLIP} is fully unsupervised, has only a minor additional computation overhead, and can be easily implemented in few lines of code. We show that \textsc{AutoCLIP} outperforms baselines across a broad range of vision-language models, datasets, and prompt templates consistently and by up to 3 percent point accuracy.

\end{abstract}

\section{Introduction}

\begin{figure}
	\begin{center}
	\includegraphics[width=.9\linewidth]{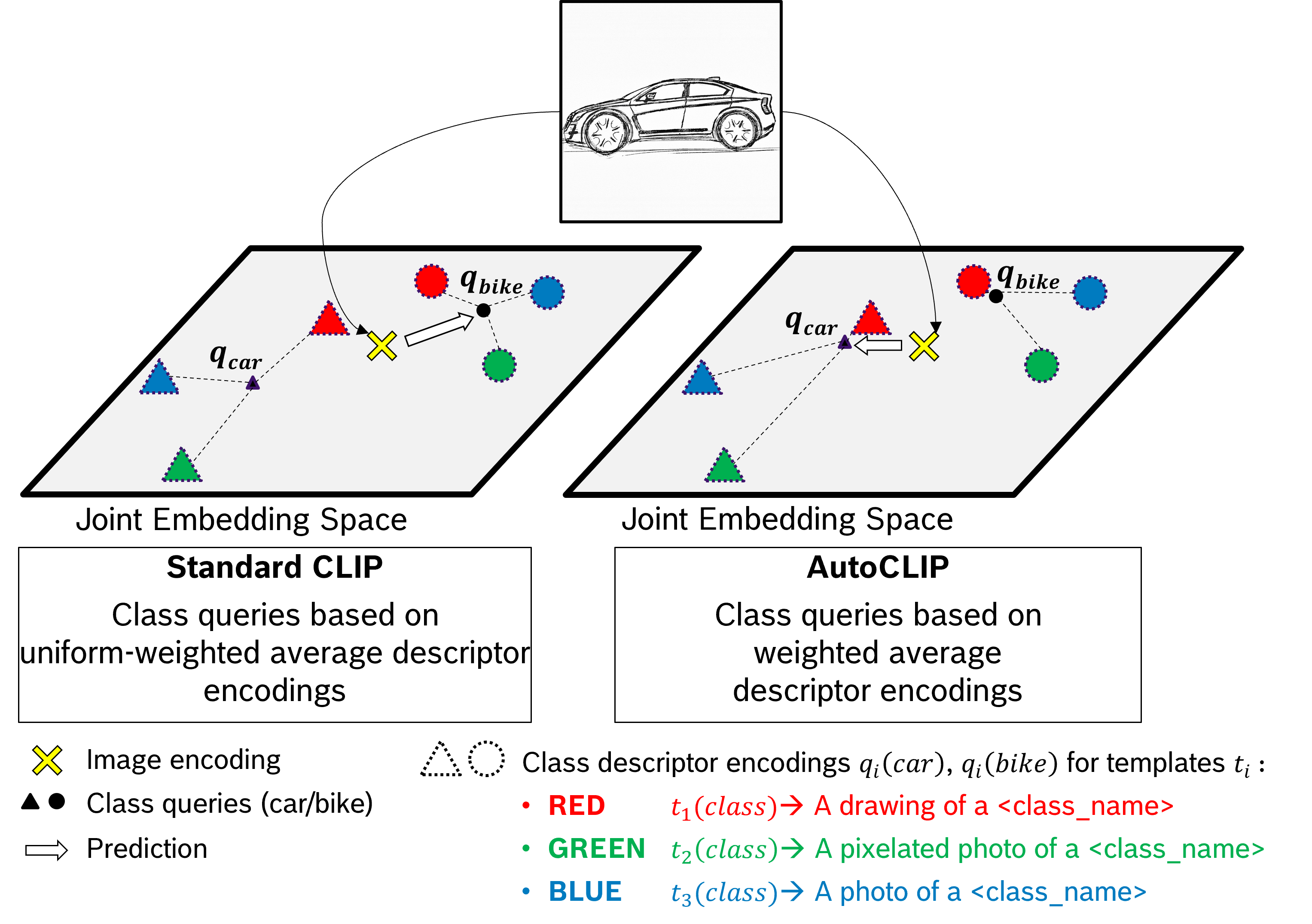} 
    \includegraphics[width=.9\linewidth]{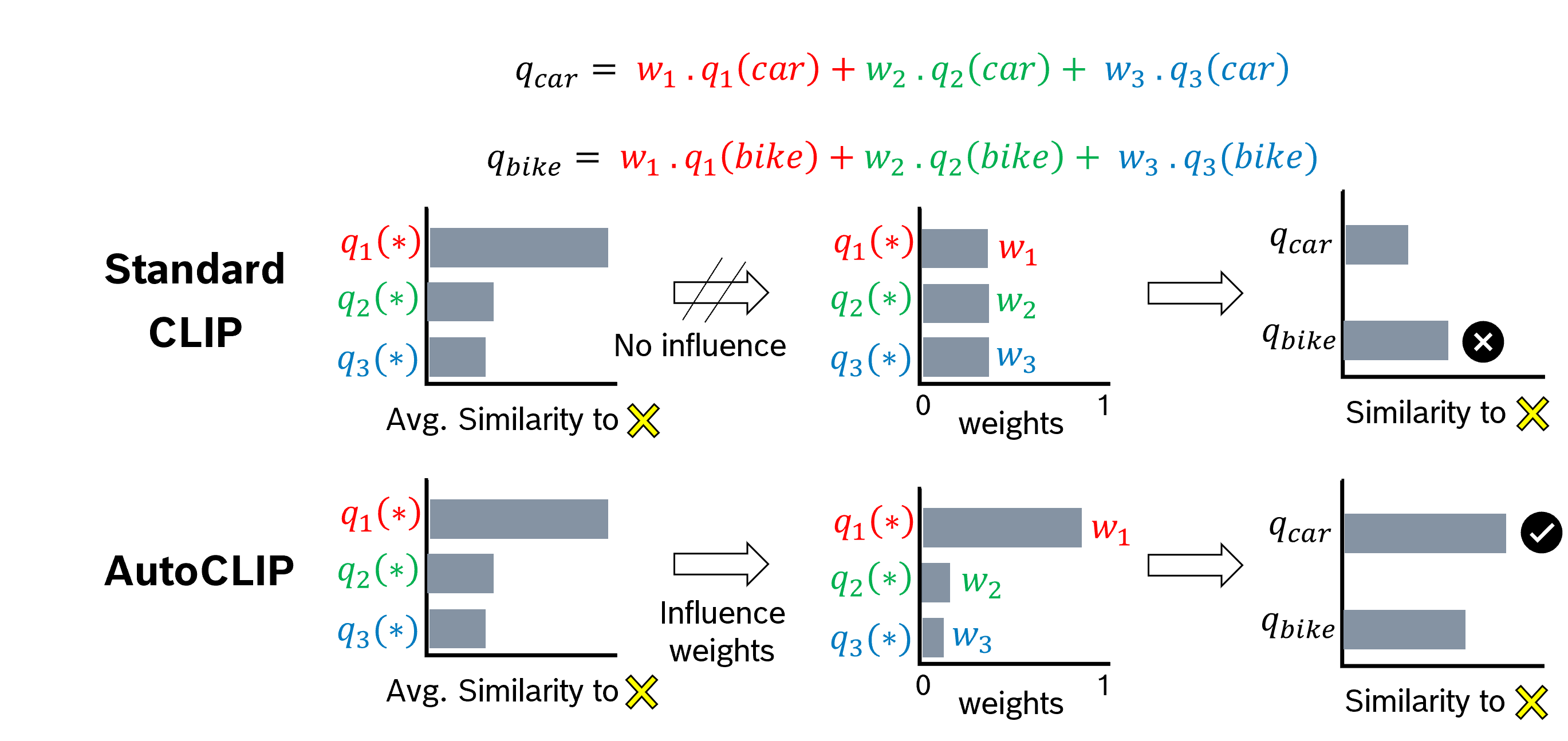} 
        \vspace{-0.25cm}
		\caption{\textbf{Conceptual Illustration of \textsc{AutoCLIP}.} CLIP's zero-shot classifiers 
        are based on a set of prompt templates $t_i$ (``A photo of a $<$class\_name$>$'', ``A drawing of a $<$class\_name$>$'', ...). Inserting class names $c$ into these templates gives a set of class descriptors that are encoded into a joint embedding space together with the respective image. Standard CLIP averages encoded class descriptors $q_i(c)$ into class queries $q_c$, 
        and classifies to the class that has maximal cosine similarity with the encoded image. However, this ignores that some prompt templates describe the image of interest better than others (their embeddings have higher average similarity): for instance, when the image is a drawing, the template ``A drawing of a $<$class\_name$>$'' results in stronger class descriptors than other templates and should thus be weighted higher when computing class queries. \textsc{AutoCLIP} determines such weights directly from class descriptor-image similarities in the embedding space. 
        Here, the car image is taken from \cite{atkinson2015car}.
        } 
		\label{Figure:autoclip_teaser}
	\end{center}
\end{figure}

Classifiers built upon vision-language models (VLMs) such as CLIP \citep{radford2021learning} and CoCa \citep{yu2022coca} have shown strong zero-shot transfer capabilities across various tasks. Such zero-shot transfer is appealing since it allows for obtaining high-performing classifiers on novel domains without the overhead of data acquisition and labelling.
However, it has been observed that prompt engineering plays a crucial role for obtaining strong zero-shot classifiers, that is: zero-shot classifiers derived from VLMs need to be constructed based on a set of prompt templates (parameterized by the class name) that cover potential variation of the domain. These prompt templates can be hand-designed \citep{radford2021learning}, generated by a large-language model \citep{menon2022visual}, or randomly generated \citep{roth2023waffling}. 

Prompts can also be learned via test-time prompt tuning (TPT) \citep{shu2022test, Zhao2023TestTimeAW}. This approach makes the zero-shot classifier adaptable to the datum of interest, which is possible by effectively leveraging the knowledge of the general-purpose VLM.
\citet{shu2022test} tune prompts so that the predictive entropy for a single image is minimized, while \citet{Zhao2023TestTimeAW} maximizes a CLIP reward. These prior TPT methods require the VLM's image encoder to process several augmentations for each image. 
Moreover, gradients with respect to the prompts require backpropagation through the VLM's text encoder, thereby substantially increasing the overall inference cost.

 We propose to not tune the prompts but instead use a large set of predefined and fixed prompt templates and to adapt the weights of those prompt templates for each image at test-time. This approach has the major advantage that adaptation takes place entirely in the embedding space without requiring additional forward or backward passes through the VLM's encoders, which significantly lowers the test-time computation and memory overhead compared to prior TPT methods. Our work is similar to \citet{Allingham2023Contrastively}, but comes with the major advantages that our approach can adapt weights for single samples and does not require access to the pre-training feature distribution.

We briefly summarize the standard way of constructing zero-shot classifiers from VLMs (see Figure \ref{Figure:autoclip_teaser} left). At first, a collection of prompt templates is instantiated for each class to form a set of class descriptors (e.g., ``A photo of a \emph{car}'', and ``A drawing of a \emph{car}'' are sample class descriptors of class \emph{car}). These descriptors are processed by the text encoder and the resulting encoded descriptors are averaged to obtain the image-independent class queries (e.g. $q_{car}$). Besides, the image encoder processes the input image to be classified to get the image encoding, which lies in the same embedding space as class queries. The cosine similarity of the encoded image to every (averaged) class query is computed, and the output prediction is assigned to the class with maximum similarity.

This work follows a similar zero-shot classification setup, except that we change how class queries are computed. Instead of a simple average of the encoded class descriptors, we propose to take a weighted average, wherein weights of the encoded class descriptors are automatically tuned for each image separately. The weights are determined in a manner that prompt templates whose resulting class descriptors are closer to the respective image embedding get higher weightage than those being less similar (see Figure \ref{Figure:autoclip_teaser} right). Our approach is motivated by the intuition that prompt templates with high similarity describe relevant properties of the image better than ones with lower similarity (see Figure \ref{Figure:weights} for evidence supporting this intuition). We denote our method that automatically adapts the weights of the encoded class descriptors for each image as \textsc{AutoCLIP}. We provide a basic implementation of \textsc{AutoCLIP} at \url{https://github.com/boschresearch/autoclip}.

We empirically show that \textsc{AutoCLIP} improves the performance of zero-shot classifiers across many datasets, VLMs, and prompt strategies with little inference-time overhead. Note that \textsc{AutoCLIP} is fully zero-shot as it does not require any supervision from the target task. Furthermore, \textsc{AutoCLIP} makes no assumptions on the underlying VLM and can thus be broadly applied, potentially also to multi-modal models beyond VLMs such as ImageBind \citep{girdhar2023imagebind}.

Overall, our main contributions are as follows: we introduce \textsc{AutoCLIP} (Section \ref{Subsection:AutoCLIP}), a novel procedure for constructing zero-shot classifiers from vision-language models. \textsc{AutoCLIP} leverages statistics of class descriptor-image similarities to automatically determine weights of the prompt templates. We further discuss a method for automatically tuning \textsc{AutoCLIP}'s step size such that the entropy of the prompt template's weights is controlled (Section \ref{Subsection:EntropyTuning}). We propose a default entropy reduction factor, which is shared across all the experiments. By this, \textsc{AutoCLIP} comes essentially without free hyperparameters, which is important as hyperparameters cannot be tuned in zero-shot settings. We evaluate \textsc{AutoCLIP} on a large number of datasets, vision-language models, and prompt templates (Section \ref{Section:Experiments}) as well as in a controlled setting (Section \ref{Section:Controlled_Setting}). We find that it improves performance on the vast majority ($85\%$) of settings, by $0.45$  percent point accuracy on average, and by up to $3$ percent point in some settings. These gains come essentially for free with the only cost being a very small inference time overhead (see Section \ref{appendix:inference_time} in the appendix), as our approach operates entirely in the embedding space. Considering these benefits, we believe that the proposed \textsc{AutoCLIP} can serve as a default zero-shot inference strategy for VLMs.


\section{Related Work}

\textbf{Vision-Language Pretraining.} Deep learning with vision-language pretraining has enabled zero-shot transfer capabilities, i.e., the resulting vision-language models (VLMs) are able to perform zero-shot classification on vastly diverse unseen target datasets given only text prompts of individual target classes. CLIP is one of the state-of-the-art VLMs pretrained on the well-curated WebImageText dataset containing 400 million image-text pairs using a contrastive loss \citep{radford2021learning}. In terms of datasets used, ALIGN requires less dataset preprocessing enabling training on a dataset of over a billion image-text pairs \citep{jia2021scaling}. Florence \citep{yuan2021florence} expands models to other common modalities (e.g., videos). In terms of the training loss, CoCa \citep{yu2022coca} leverages an additional captioning loss allowing models to be used in generative applications. In our work, we study how to optimally use text prompts of the target classes with these VLMs.

\textbf{Prompt Construction.} Conventionally, one or several manually designed text prompts per target class are employed for zero-shot classification \citep{radford2021learning,jia2021scaling}. Recent research demonstrates that introducing additional prompts can improve overall performance. DCLIP \citep{menon2022visual} generates additional prompts based on querying the large-language model GPT-3 \citep{brown2020language}.  WaffleCLIP \citep{roth2023waffling} has shown that classification performance can be further boosted by appending random words or characters to predefined prompt templates. To derive a zero-shot classifier, these works weight all text prompts uniformly. In contrast, we propose an approach to adjust weights of individual prompts per input sample dynamically at test time.

\textbf{Test-Time Adaptation.} Our work can be considered as a test-time adaption approach for VLMs. TENT \citep{wang2020tent} demonstrates that adapting models to minimize prediction entropy can improve model performance at test time. In the context of VLMs, TPT \citep{shu2022test} optimizes prompts of target classes based on the entropy minimization objective. RLCF \citep{Zhao2023TestTimeAW} demonstrates that minimizing the entropy objective can lead to overfitting under distribution shift and proposes adaptation based on average CLIP scores. In contrast to these previous works, we do not perform any adaptation of prompts or model parameters, but refine weights of individual (encoded) prompts, which is considerably cheaper in terms of computation and memory consumption. Most similar to our work is Zero-shot Prompt Ensembling (ZPE) \citep{Allingham2023Contrastively}, which also determines prompt weights in embedding space. However, ZPE requires an entire batch of target domain samples and the availability of image features representing the feature distribution in pre-training (``source domain''). In contrast, our work operates on single images in a source-free setting.


\section{AutoCLIP}
We outline the common approach for building zero-shot classifiers for VLMs like CLIP in Section \ref{Subsection:BackgroundZeroShot}. Thereupon, we detail our proposed \textsc{AutoCLIP} as an auto-tuned alternative in Section \ref{Subsection:AutoCLIP}, followed by describing how the required gradient can be calculated in closed-form in Section \ref{Subsection:GradientComputation}, and finally explain how \textsc{AutoCLIP}'s step size can be automatically determined in Section \ref{Subsection:EntropyTuning}.

\subsection{Background: Zero-Shot Classifiers for Vision-Language Models} \label{Subsection:BackgroundZeroShot}
Let us consider a classification task $\mathcal{X} \mapsto \mathcal{C}$, where $\mathcal{X}$ corresponds to the input domain and $\mathcal{C}=\{c_1, \dots, c_C\}$ is a set of $C$ classes. We assume that there exists a pretrained VLM such as CLIP that provides a joint embedding space $\mathcal{E}$ and corresponding embedding functions $E_X: \mathcal{X} \mapsto \mathcal{E}$  that maps input data $x \in \mathcal{X}$ into embedding space $\mathcal{E}$ and $E_T: \mathcal{T} \mapsto \mathcal{E}$ that maps text into the same embedding space $\mathcal{E}$. Let there be $K$ prompt templates $t_1, \dots t_K: \mathcal{C} \mapsto \mathcal{D}$ that map class name $c \in \mathcal{C}$ to (textual) class descriptors $d \in \mathcal{T}$. These prompt templates can be either manually designed \citep{radford2021learning}, generated by a large language model \citep{menon2022visual}, or randomly generated \citep{roth2023waffling}. 
Algorithm \ref{alg:zero_shot} summarizes the standard zero-shot classifier for VLMs: average the class descriptor encodings $e^{(d)}$ into class queries $q_j$, then compute cosine similarities $s_j$ between class query and encoded image $e^{(x)}$, and classify to the class that maximizes similarity.

\begin{algorithm}[tb]
\caption{Zero-Shot Classifier for a single sample $x$}\label{alg:zero_shot}
\begin{algorithmic}[1]
\State \LeftComment{0}{Generate $K \times C$ class descriptors}
\State $d \gets \{t_i(c_j) \,\vert\, i \in \{1,\dots,K\}, j \in \{1,\dots,C\}\}$ 
\State \LeftComment{0}{Encode image of interest $x$ with VLM}
\State $e^{(x)} \gets E_X(x) / \vert\vert E_X(x) \vert\vert_2$ 
\State \LeftComment{0}{Encode all class descriptors with VLM}
\State $e^{(d)}_{ij} \gets E_T(d_{ij}) / \vert\vert E_T(d_{ij}) \vert\vert_2$ 
\color{red}
\State $w_i \gets 1 / K$ \Comment{Uniform prompt template weights}
\color{black}
\For{$j \in 1,\dots,C$}
   \State \LeftComment{0}{Class queries as average class descriptor encodings}
   \State $q_j \gets \sum_{i=1}^K w_i e^{(d)}_{ij}$   
   \State \LeftComment{0}{Cosine similarity between $e^{(x)}$ and class query $q_j$}
   \State $s_j \gets e^{(x)} \cdot q_j$ 
\EndFor
\State \LeftComment{0}{Assign $x$ to class $c_{j^\star}$ with maximum similarity}
\State $j^\star \gets \arg\max_j s_j$
\end{algorithmic}
\end{algorithm}

\subsection{Auto-Tuning Zero-Shot Classfiers} \label{Subsection:AutoCLIP}
\textsc{AutoCLIP} modifies Line 7 in Algorithm \ref{alg:zero_shot}. Instead of computing class queries as simple average of class descriptor encodings $q_j = 1 / K \sum_{i=1}^K e^{(d)}_{ij}$, \textsc{AutoCLIP} uses a weighted average: $q_j =\sum_{i=1}^K w_i e^{(d)}_{ij}$ with tunable $w$, satisfying $w_i \geq 0,\; \sum_{i=1}^K w_i = 1$, which we enforce by reparameterizing $w = \softmax(\rho)$ and $\rho \in \mathbb{R}^K$. \textsc{AutoCLIP}'s guiding intuition (see Figure \ref{Figure:autoclip_teaser}) is to assign higher weights $w_i$ to prompt templates $t_i$ that result in class descriptor encodings $e^{(d)}_{ij}$ that are more similar to the encoded image $e^{(x)}$, that is: $t_i$ with large $e^{(xd)}_{ij} = e^{(d)}_{ij} \cdot e^{(x)}$ ($j=1,\dots, C$). This is inspired by the observation that class descriptors having higher similarity in the embedding space describe the image better (according to contrastive pretraining objectives in typical VLMs). In practice, \textsc{AutoCLIP} tunes $w$ on a per-sample basis by one step of gradient ascent on a $\logsumexp$-based objective function, which we detail below.

When determining the template's weights $w$, we have $C$ descriptor-image similarities $e^{(xd)}_{ij}$ for each template $t_i$. AutoCLIP needs to aggregate those $C$ similarities across classes when assigning larger weights to more relevant prompt templates. Intuitively, simply averaging all $C$ similarities (``mean'' aggregation) ignores that, in the classification objective, we ultimately only care about classes that result in the descriptors closest to $e^{(x)}$; however, taking only the class with highest similarity per template into account (``max'' aggregation) ignores inherent ambiguity in the image and was found to be suboptimal \citep{roth2023waffling}. We propose a middle ground of aggregating via a  smooth approximation to the maximum function via $\logsumexp_j(e^{(xd)}_{ij}) = \log \sum_{j=1}^C \exp e^{(xd)}_{ij}$. This $\logsumexp$ aggregation takes all classes into account but assigns higher importance to more relevant classes (ones resulting in higher similarities to the image $x$). \textsc{AutoCLIP} then determines weights $w_i$ such that $\logsumexp_j(s_j) = \logsumexp_j(\sum_{i=1}^K w_i e^{(xd)}_{ij}) =\logsumexp_j(\softmax(\rho)  \cdot e^{(xd)}_{:j})$ gets increased by one step of gradient ascent in the direction of $\nabla_\rho \logsumexp_j(\softmax(\rho) \cdot e^{(xd)}_{:j})$. We note that $-\logsumexp$ has been interpreted as the energy function of a data point (for appropriately trained classifiers) \citep{Grathwohl2020Your}; in this view, \textsc{AutoCLIP} can be interpreted as minimizing the energy and maximizing the probability density $p(x)$ of $x$ under the zero-shot classifier (see Section \ref{appendix:logsumexp} for more details).

We summarize \textsc{AutoCLIP} in Algorithm \ref{alg:auto_clip}.
We initialize $\rho = \mathbf{0}$, which corresponds to an unweighted average of the class descriptor encodings (Line 8). Similar to Algorithm \ref{alg:zero_shot}, we compute the pairwise cosine similarities $s_j$ between encoded image $e^{(x)}$ and class queries $q_j$ (Line 9-14). Instead of directly classifying to the class with maximum similarity to the image, \textsc{AutoCLIP} updates the class descriptor weights first. For this, the gradient $g=\nabla_\rho \logsumexp_j(s_j)$ is computed (Line 16), an appropriate step size $\alpha$ is selected (Line 18, see Section \ref{Subsection:EntropyTuning}), and $\rho = \alpha \cdot g$ and $w= \softmax(\rho)$ are updated (Line 20). Based on the new $w$, \textsc{AutoCLIP} computes updated class queries $q_j$ and class-image similarities (Line 21-26) and finally selects the class with maximum similarity for the image (Line 28). It is worth emphasizing that \textsc{AutoCLIP} is permutation-invariant in the prompt templates $t_i$.

We note that Line 9-20 could be repeated for several iterations with smaller step sizes; however preliminary experiments indicate no advantage of doing more than one iteration. We call \textsc{AutoCLIP} ``auto-tuned'' because its  weights $w$ are automatically adapted for every input independently. Moreover, we note that in practice, models like CLIP scale $e^{(xd)}$ by a learned temperature (exponential logit scale) $\tau$ to obtain well calibrated classifiers; we use the same temperature for scaling $e^{(xd)}$ in the $\logsumexp$ aggregation (as there is no labelled data in a zero-shot setting on which a temperature could be tuned). 

\begin{algorithm}[tb]
\caption{\textsc{AutoCLIP}: Auto-Tuned Zero-Shot Classifier for a single sample $x$}\label{alg:auto_clip}
\begin{algorithmic}[1]
\State \LeftComment{0}{Generate $K \times C$ class descriptors}
\State $d \gets \{t_i(c_j) \,\vert\, i \in \{1,\dots,K\}, j \in \{1,\dots,C\}\}$ 
\State \LeftComment{0}{Encode image of interest $x$ with VLM}
\State $e^{(x)} \gets E_X(x) / \vert\vert E_X(x) \vert\vert_2$
\State \LeftComment{0}{Encode all class descriptors with VLM}
\State $e^{(d)}_{ij} \gets E_T(d_{ij}) / \vert\vert E_T(d_{ij}) \vert\vert_2$ 
\color{red}
\State \LeftComment{0}{Uniform weights $w_i = 1 / K$}
\State $\rho \gets \mathbf{0}; \quad w_i \gets \softmax(\rho)$
\color{black}
\For{$j \in 1,\dots,C$}
   \State \LeftComment{0}{Class queries as average class descriptor encodings}
   \State $q_j \gets \sum_{i=1}^K w_i e^{(d)}_{ij}$   
   \State \LeftComment{0}{Cosine similarity between $e^{(x)}$ and class query $q_j$}
   \State $s_j \gets e^{(x)} \cdot q_j$
\EndFor
\color{red}
\State \LeftComment{0}{Compute gradient (Section \ref{Subsection:GradientComputation})}
\State $g \gets \nabla_{\rho} \log \sum_{j=1}^C \exp(s_j)$
\State \LeftComment{0}{Determine stepsize (Section \ref{Subsection:EntropyTuning})}
\State $\alpha \gets \text{BISECT}(\softmaxentropy(\alpha \cdot g) - \beta\log_2 K, 0, 10^{10})$ 
\State \LeftComment{0}{Update $\rho$ with one gradient ascent step and step size $\alpha$}
\State $\rho \gets \alpha \cdot g; \quad w_i \gets \softmax(\rho)$ 
\color{black}
\For{$j \in 1,\dots,C$}
   \State \LeftComment{0}{Class queries as average class descriptor encodings}
   \State $q_j \gets \sum_{i=1}^K w_i e^{(d)}_{ij}$   
   \State \LeftComment{0}{Cosine similarity between $e^{(x)}$ and class query $q_j$}
   \State $s_j \gets e^{(x)} \cdot q_j$
\EndFor
\State \LeftComment{0}{Assign $x$ to class $c_{j^\star}$ with maximum similarity}
\State $j^\star \gets \arg\max_j s_j$ 
\end{algorithmic}
\end{algorithm}

\subsection{Closed-form Computation of Gradient}
\label{Subsection:GradientComputation}
While $\nabla_\rho \logsumexp(s)$ can be easily computed using automatic differentiation, we note that there can be runtime environments for inference such as on edge devices where running automatic differentiation is undesirable. For such cases, the gradient $\nabla_\rho \logsumexp_j(s_j)$ can also be computed in closed-form:
$\left( \nabla_\rho \logsumexp_j(s_j)\right)_i =  \sum_{k=1}^K (\sum_{j=1}^C \softmax(s)_j \cdot  e^{(xd)}_{ij}) \cdot w_i(\delta_{ik} - w_k),$
with $\delta_{ij}$ being the Kronecker delta function with $\delta_{ii} = 1$ and $\delta_{ij} = 0$ for $i \neq j$.

\subsection{Auto-Tuning the Step Size} \label{Subsection:EntropyTuning}
The only free hyperparameter of \textsc{AutoCLIP} is the step size $\alpha$. We note that in a zero-shot setting, there is by definition no labeled data on which such free hyperparameters can be tuned. Because of this, free hyperparameters need to be selected globally in a dataset-independent manner. However, a global choice for the step size $\alpha$ is problematic since the scale of the gradient $g = \nabla_\rho \logsumexp(s)$ depends on the dataset, and the step size would have to be adapted accordingly. We address this by proposing a different parameterization in which the free hyperparameter is easily interpreted and the step size $\alpha$ is a derived quantity. Specifically, we control the entropy of the query weights $w$, $\entropy(w) = -\sum_{i=1}^K w_i \log_2 w_i$. The standard, uniform weights have maximum entropy $\log_2 K$ and we set the target entropy to $\beta \cdot \log_2 K$, where the entropy reduction factor $\beta \in [0, 1]$ is the new free hyperparameter that we set globally to $\beta=0.85$. Intuitively, $\beta \to 1$ corresponds to more equally weighted prompt templates while $\beta \to 0$ to selecting the prompt template with maximum similarity. We present an ablation of the effect of $\beta$'s choice on \textsc{AutoCLIP} in Figure \ref{Figure:results_entropy_rate}.

With $\softmaxentropy(\alpha\cdot g)$ denoting the entropy of the weights $w = \softmax(\alpha\cdot g)$, selecting the step size $\alpha$ is now equivalent to solving for $f(\alpha) = 0$ for $f(\alpha) = \softmaxentropy(\alpha\cdot g) - \beta \cdot \log_2 K$. As $\softmaxentropy(\alpha\cdot g)$ monotonically decreases with $\alpha$, we use bisection on $\alpha \in [0, 10^{10}]$ for  finding $\alpha$ with $f(\alpha) \approx 0$. We note that $\softmaxentropy(0\cdot g) = \log_2 K$ and thus $f(0) > 0$ for all $\beta < 1$; similarly, $\softmaxentropy(\alpha\cdot g) \approx 0$ for $\alpha=10^{10}$ in all settings we considered and thus $f(10^{10}) < 0$ for all $\beta > 0$, which together satisfies the prerequisites for running bisection. The additional bisection has little overhead compared to the  cost of encoding the image $x$ with $E_x$ (see Section \ref{appendix:inference_time} in the appendix for details).

\begin{figure*}[p]
	\begin{center}
	\includegraphics[width=\linewidth]{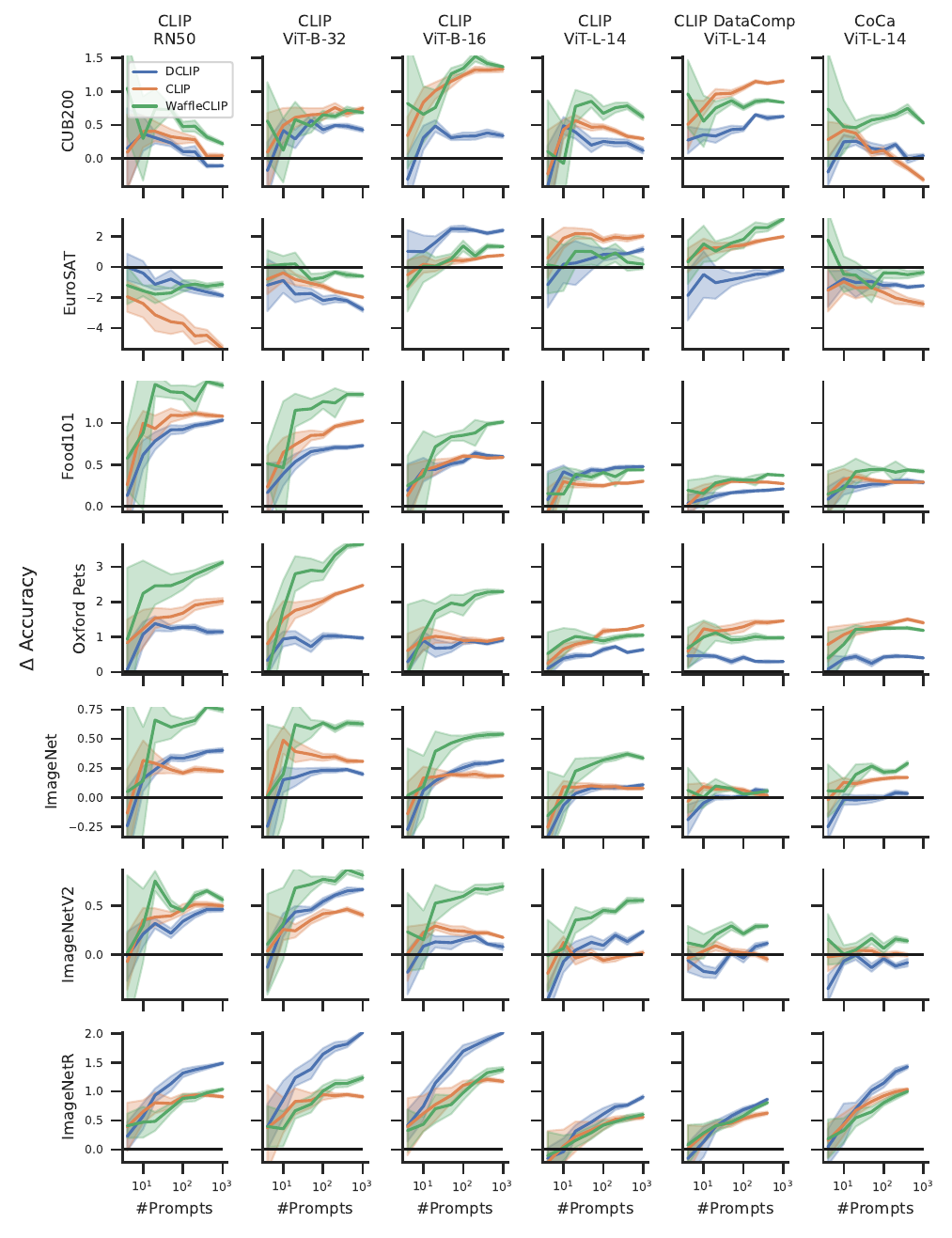} 
        \vspace{-1cm}
		\caption{Accuracy improvement ($\Delta$ Accuracy) of \textsc{AutoCLIP} over baseline zero-shot classifier across models, datasets, and prompt ensembles. Shown are mean and standard error over 7 runs.}
		\label{Figure:results_main}
	\end{center}
\end{figure*}

\begin{figure*}[tb]
	\begin{center}
    \includegraphics[width=\linewidth]{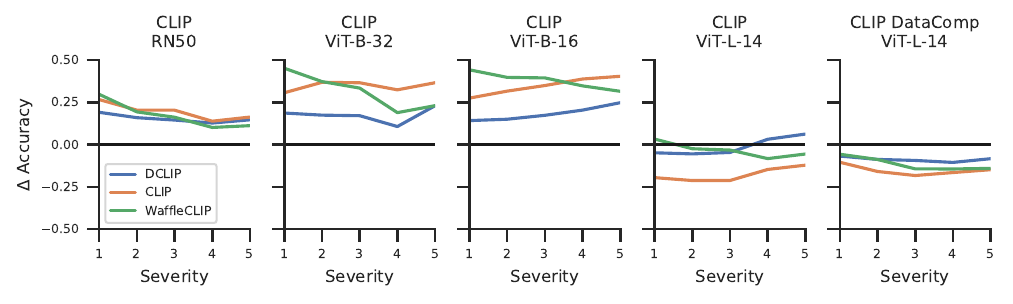} 
        \vspace{-.5cm}
		\caption{ImageNet-C accuracy improvement ($\Delta$ Accuracy) of \textsc{AutoCLIP} over baseline zero-shot classifier for $K=100$ across models, corruption severity and prompt ensembles, averaged over corruptions and 7 runs.}
		\label{Figure:results_imagenetc}
	\end{center}
\end{figure*}

\section{Experiments} \label{Section:Experiments}
\paragraph{Experimental Setting}
In this section, we compare \textsc{AutoCLIP} to standard zero-shot classifiers on a wide range of zero-shot image classification benchmarks and a variety of settings. We conduct experiments on the datasets CUB200 \citep{welinder2010caltech}, EuroSAT \citep{helber2019eurosat}, Food101 \citep{bossard2014food}, Oxford Pets \citep{parkhi2012cats}, ImageNet \citep{russakovsky2015imagenet}, ImageNetV2 \citep{kornblith2019better}, ImageNet-R \citep{hendrycks2021many}, and ImageNet-C \citep{hendrycks2019robustness}. We study six different vision-language models: from CLIP \citep{radford2021learning}, we use ResNet-50 (RN50) \citep{He2015DeepRL} and vision transformer (ViT-B/32, ViT-B/16, and ViT-L/14) model variants \citep{dosovitskiy2021an}. Moreover, we use the ViT-L/14 model variant from DataComp \citep{gadre2023datacomp} and the one trained with CoCa \citep{yu2022coca}.

Additionally, we study three ways of generating prompt templates: 1) using the 80 manually designed templates from \citet{radford2021learning} (CLIP), 2) templates based on querying a large-language model (DCLIP) \citep{menon2022visual}, and 3) templates that append random words or characters to predefined prompt templates (WaffleCLIP) \citep{roth2023waffling}. We vary the number of templates from $K=4$ to $K=500$; if there is a fixed number of templates available such as in CLIP/DCLIP, templates are sampled with replacement. To account for randomness in the template construction/sampling, we report results averaged over $7$ runs. We base our implementation on \url{https://github.com/ExplainableML/WaffleCLIP} from \cite{roth2023waffling} and highly appreciate their code release under a permissible license. We report the difference of accuracy of \textsc{AutoCLIP} compared to the baseline zero-shot classifier with uniform prompt template weights ("$\Delta$ Accuracy"). Absolute performance across different datasets and VLMs is shown in Table \ref{Table:results_main_waffle} (and in Table \ref{Table:results_main_dclip} and Table \ref{Table:results_main_clip} in the appendix).

\begin{table*}[tb]
\begin{center}
\begin{tabular}{lcccccc}
\toprule
&CLIP & CLIP & CLIP & CLIP & DataComp & CoCa\\
&RN50 & ViT-B-32 & ViT-B-16 & ViT-L-14 & ViT-L-14 & ViT-L-14\\
\midrule
              CUB200  & 47.75 (+0.5) & 52.84 (+0.7) & 57.12 (+1.3) & 64.43 (+0.7) & 84.79 (+0.8) & 73.90 (+0.6) \\
             EuroSAT  & 34.95 (-1.2) & 46.16 (-0.7) & 55.93 (+1.4) & 55.09 (+0.6) & 65.09 (+1.8) & 54.77 (-0.4) \\
             Food101  & 80.26 (+1.4) & 84.13 (+1.3) & 88.85 (+0.9) & 93.71 (+0.4) & 94.52 (+0.3) & 90.46 (+0.4) \\
         Oxford Pets  & 83.09 (+2.6) & 85.63 (+2.9) & 85.89 (+1.9) & 91.64 (+0.9) & 92.82 (+0.9) & 92.03 (+1.2) \\
            ImageNet  & 60.42 (+0.6) & 63.80 (+0.6) & 68.70 (+0.5) & 75.89 (+0.3) & 79.07 (+0.0) & 75.63 (+0.2) \\
          ImageNetV2  & 53.44 (+0.4) & 56.49 (+0.8) & 62.54 (+0.6) & 70.17 (+0.4) & 72.21 (+0.2) & 68.08 (+0.1) \\
           ImageNetR  & 29.32 (+0.9) & 51.04 (+1.0) & 59.13 (+1.0) & 73.98 (+0.4) & 78.85 (+0.6) & 75.59 (+0.8) \\
\bottomrule
\end{tabular}
		\caption{Accuracy of \textsc{AutoCLIP} (and $\Delta$ Accuracy  to baseline zero-shot classifier in parenthesis) for $K=100$ WaffleCLIP prompt templates across models and datasets,  averaged over 7 runs.}
		\label{Table:results_main_waffle}
\end{center}
\end{table*}

\begin{figure}[tb]
	\begin{center}
	\includegraphics[width=.9\linewidth]{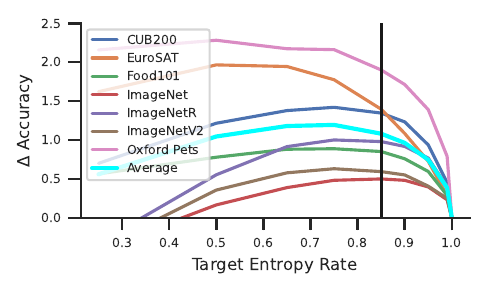} 
    \vspace{-.5cm}
		\caption{Ablation on target entropy rate $\beta$. Shown is the
accuracy improvement ($\Delta$ Accuracy) of \textsc{AutoCLIP} over baseline zero-shot classifier for a CLIP ViT-B-16, and 100 WaffleCLIP prompt templates, averaged over 7 runs.}
		\label{Figure:results_entropy_rate}
	\end{center}
\end{figure}

\paragraph{Results}
We present the main results in Figure \ref{Figure:results_main}. Overall, the figure contains $990$ different combinations comparing \textsc{AutoCLIP} with the baseline; \textsc{AutoCLIP} is better in $840$ cases ($\approx 85\%$) and on average it is better by $0.45$ percent point accuracy. We also observe a trend that for larger number of prompt templates $K$, the advantage of \textsc{AutoCLIP} ($\Delta$ Accuracy averaged across datasets, models and CLIP/DCLIP/WaffleCLIP) increases: from $\Delta=0.06$ for $K=4$ over $\Delta=0.33$ for $K=10$ and $\Delta=0.49$ for $K=50$ to $\Delta=0.57$ for $K=200$. When aggregating over models, datasets and number of prompt templates, \textsc{AutoCLIP} achieves the largest average improvement for WaffleCLIP ($\Delta=0.61$), but still improves for CLIP ($\Delta=0.40$) and DCLIP ($\Delta=0.29$). Taken together,  the findings indicate that \textsc{AutoCLIP} benefits from larger (increased K) and more diverse (WaffleCLIP) sets of prompt templates. 

When comparing different vision-language models, \textsc{AutoCLIP} brings the biggest benefit for CLIP ViT-B-16 ($\Delta=0.68$) and the smallest one for CoCa ViT-L-14 ($\Delta=0.19$), with all other models having average $\Delta$ between $0.36$ and $0.52$. Comparing different datasets, \textsc{AutoCLIP} performs strongest on Oxford Pets ($\Delta=1.15$) and worst on EuroSAT ($\Delta=-0.24$); we hypothesize that this is because EuroSAT is in general a challenging dataset for CLIP on which the image encoder produces embeddings that are not very informative about image properties, which deteriorates the prompt weight selection as it becomes harder to decide which prompt describes an image of interest well. We note that EuroSAT is the only setting on which \textsc{AutoCLIP} hurts performance on average; on all other datasets, \textsc{AutoCLIP} improves performance: $\Delta(\text{CUB200})=0.5$, $\Delta(\text{Food101})=0.52$, $\Delta(\text{ImageNet})=0.17$, $\Delta(\text{ImageNetV2})=0.2$, and $\Delta(\text{ImageNetR})=0.71$. While these improvements are of modest magnitude, they essentially come for free by a mere change of the inference procedure.

In Figure \ref{Figure:results_imagenetc}, we present results on ImageNet-C for WaffleCLIP with $K=100$ for different severities and averaged across corruptions. \textsc{AutoCLIP} consistently improves performance for the smaller vision-language models (RN50, ViT-B-32, ViT-B-16) and sees a minor drop of performance for the two ViT-L-14 variants. Averaged across all models, corruptions, and severities, \textsc{AutoCLIP} improves performance by $\Delta=0.11$. We provide plots for each corruption separately for WaffleCLIP prompt templates in the appendix in Figure \ref{Figure:imagenetc_detailed}. The biggest average benefit of \textsc{AutoCLIP} is obtained for the low-frequency corruptions  ``saturate'' ($\Delta=0.22$), ``brightness'' ($\Delta=0.22$), and ``contrast'' ($\Delta=0.23$); the smallest average benefit for ``shot-noise'' ($\Delta=0.05$) and ``snow'' ($\Delta=0.06$). 

\paragraph{Ablations}
We ablate \textsc{AutoCLIP}'s choice of the target entropy rate $\beta$ (which defaults to 0.85) and the objective function (defaults to $\logsumexp$). In Figure \ref{Figure:results_entropy_rate}, we observe that \textsc{AutoCLIP}'s performance for most datasets does not depend strongly on the specific choice of the target entropy rate $\beta$ as $\Delta$ Accuracy stays relatively constant in the range $\beta \in [0.7, 0.9]$. This is a desirable property as in a zero-shot setting without labeled data, tuning $\beta$ per dataset would be infeasible. For two datasets (Oxfort Pets and EuroSAT), our default value of $\beta=0.85$ was suboptimal and a considerably smaller choice of $\beta=0.7$ would have obtained considerably better results. Also on average, $\beta=0.7$ performs favorably and we recommend this choice for future work on other datasets and tasks. We provide results for a similar experiment in which we directly control the step size $\alpha$ in Section \ref{appendix:step_size_selection}. Directly controlling $\alpha$ reduces computation overhead further, but optimal choices of step size $\alpha$ vary more strongly across datasets than choices for the target entropy rate $\beta$.

\begin{figure*}[tb]
	\begin{center}
	\includegraphics[width=\linewidth]{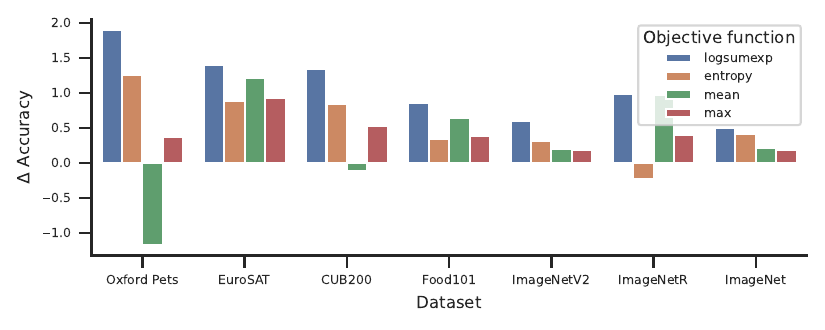} 
    \vspace{-.5cm}
		\caption{Comparison of different objective functions for auto-tuning. Shown is the accuracy improvement ($\Delta$ Accuracy) of \textsc{AutoCLIP} over baseline zero-shot classifier for a ViT-B-16, and 100 WaffleCLIP prompt templates, averaged over 7 runs.}
		\label{Figure:results_tta_losses}
	\end{center}
\end{figure*}

We motivated the choice of $\logsumexp$ as \textsc{AutoCLIP}'s aggregation/objective function in Section \ref{Subsection:AutoCLIP} as striking a good compromise between max and mean aggregation. In Figure \ref{Figure:results_tta_losses}, we empirically confirm that the $\logsumexp$ aggregation performs favorably compared to max/mean aggregation on all datasets. Moreover, it also outperforms entropy aggregation, which is a popular choice for test-time adaptation \citep{wang2020tent,shu2022test}.

\begin{figure*}[htb!]
	\begin{center}
	\includegraphics[width=\linewidth]{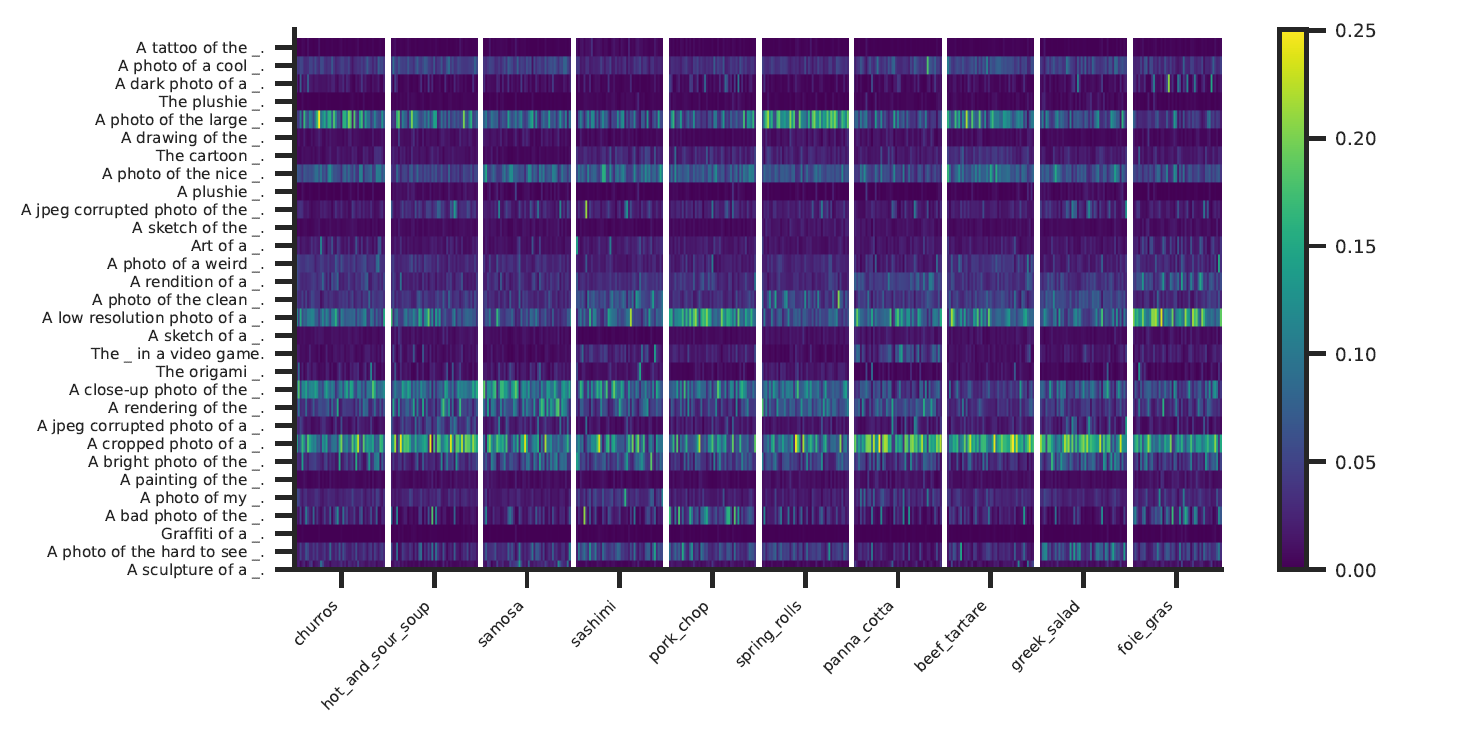} 
    \vspace*{-.75cm}
		\caption{Illustration of prompt template weights $w$ on 500 samples from the Food101 dataset, with blocks of 50 samples belonging to the same (unknown) class. CLIP backbone is a ViT-B-16 and 30 CLIP prompt templates are used.}
		\label{Figure:weights}
	\end{center}
\end{figure*}

In Figure \ref{Figure:weights}, we show the prompt template weights ($K=30$) obtained by \textsc{AutoCLIP} on 500 Food101 samples. Samples are structured in 10 blocks of 50 samples each, where each block corresponds to one class. Prompt template weights are relatively similar for instances belonging to the same (unknown) class but vary across classes. Some templates like the ones starting with ``A tattoo of...'' or ''A drawing of...'' get consistently low weights as the images of the Food101 dataset do not look like tattoos or origami, while templates starting with ``A photo of...'' tend to get higher weights, as Food101 contains mostly actual photos. Note that the weight distribution looks different on other datasets like ImageNet-R, with higher weights for ``artistic'' prompts (see Figure \ref{Figure:weights_imagenetr} in the appendix). Overall, this confirms that \textsc{AutoCLIP} can adapt the zero-shot classifier on the fly to properties of the respective image. Moreover, \textsc{AutoCLIP} provides accuracy improvements with only minor  additional inference overhead 
as discussed in Section \ref{appendix:inference_time} in the appendix.

\section{Analysis in a Controlled Setting} \label{Section:Controlled_Setting}
We study \textsc{AutoCLIP} in a controlled setting, in which we directly sample embedding vectors in an embedding space corresponding to encoded image and class descriptors, without actually encoding images or text prompts. By this, we can control key properties of the embeddings and study how they influence \textsc{AutoCLIP}'s performance. While the setting is strongly simplified, we will nevertheless gain some insights that provide possible explanations for some of the key findings from Section \ref{Section:Experiments}. Code for reproducing the results of this section is available at \url{https://github.com/boschresearch/autoclip}.

We set the number of classes to $C=5$, the number of embedding dimensions to $d=128$, the number of prompt templates to $K=10$, and the number of instance to $200$. We sample class descriptor embeddings $E_T(d_{ij})$ as follows: let $c_j \sim \mathcal{N}(0, 1, d) \in \mathbb{R}^d$ be $C$ $d$-dimensional standard normal-distributed class means, let $p_i \sim \mathcal{N}(0, 1, d) \in \mathbb{R}^d$ be $K$ $d$-dimensional standard normal-distributed prompt embedding means, and $\Psi_{ij} \sim \mathcal{N}(0, 1, d) \in \mathbb{R}^d$ be $KC$ $d$-dimensional standard normal-distributed prompt-class coupling terms. We then set $E_T(d_{ij}) = (1 - \rho) (c_j + p_i) + \rho \Psi_{ij}$, where $\rho \in [0, 1]$ controls the ``entanglement'' between class and prompt template embeddings. Intuitively, $\rho=0$ simulates a setting in which the VLM's text encoder perfectly separates class and prompt template related information such that their combination is additive. Increasing $\rho$ results in a stronger entanglement such that class and prompt template-related information interact more strongly and are no longer additive. We then set the image embeddings  corresponding to a $E_T(d_{ij})$ to $E_X(x) = E_T(d_{ij}) + \xi$ with $\xi \sim \mathcal{N}(0, \varepsilon, d)$. Here $\varepsilon$ controls the ``instance noise'', that is: how much the image embeddings are spread around the corresponding class descriptor embedding.

\begin{figure*}[tb]
	\begin{center}
	\includegraphics[width=\linewidth]{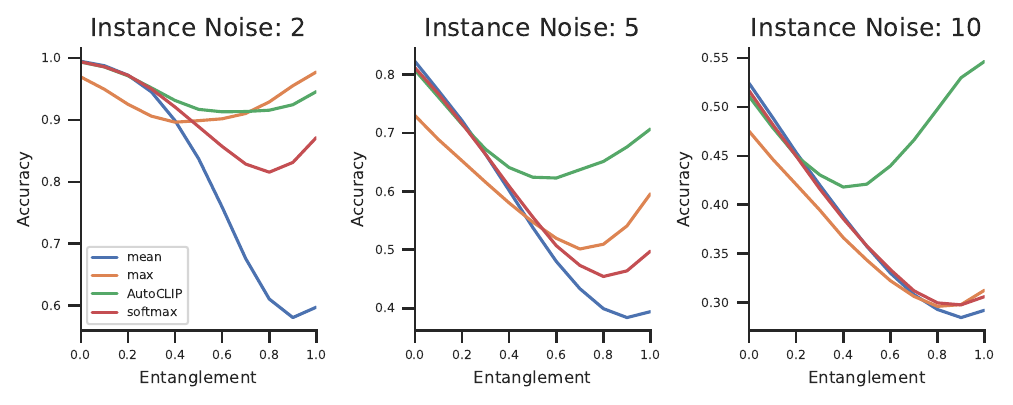} 
    \vspace{-.5cm}
		\caption{Comparison of \textsc{AutoCLIP} to ``mean'' aggregation (Line 10 in Algorithm \ref{alg:zero_shot}) and ``max'' aggregation ($s_j \gets \max_i e_{ij}^{(d)} \cdot q_j$) and ``softmax'' aggregation (which determines weights based on the respective image-prompt template similarity, see Section \ref{sec:softmax_aggr}) in a controlled and simplified setting. ``Instance Noise'' controls how strongly instance encodings vary from their respective mean. ``Entanglement'' controls how strongly the text encoding of prompt template and class name are entangled. Shown is mean over 100 random seeds.}
		\label{Figure:results_toy_data}
	\end{center}
\end{figure*}

Figure \ref{Figure:results_toy_data} compares the accuracy of mean-, max- and AutoCLIP-aggregation ($\beta=0.85$) for different values of entanglement $\rho$ and instance noise $\varepsilon$.  In general, lower entanglement favors mean aggregation (the standard in CLIP), while higher entanglement favors max-aggregation. Usually, VLM text encoders are good in terms of disentangling concepts, which explains why prior work \citep{roth2023waffling} has found max-aggregation to perform inferior compared to mean-aggregation. On the other hand, we observe that  \textsc{AutoCLIP} nearly always outperforms max-aggregation (except for the small instance noise and strong entanglement setting), also outperforms mean-aggregation for moderate entanglement ($\rho > 0.4$), and performs very close to mean-aggregation for smaller entanglement. 

This provides a possible explanation for the findings from Figure \ref{Figure:results_main}: for smaller (and weaker) VLMs, the text embeddings are more entangled and thus \textsc{AutoCLIP} provides stronger benefits compared to standard mean-aggregation. For larger VLMs like ViT-L-14 based ones, the entanglement decreases and thus the benefit of \textsc{AutoCLIP} is smaller. Moreover, for small entanglement and large instance noise, \textsc{AutoCLIP} can also be slightly worse than mean-aggregation in Figure \ref{Figure:results_toy_data}. This settings likely corresponds to the ViT-L-14 (low class-prompt entanglement) on ImageNet-C (large instance noise) in Figure \ref{Figure:results_imagenetc}, where \textsc{AutoCLIP} also performs slightly worse than mean-aggregation. However, in general Figure \ref{Figure:results_toy_data} indicates that \textsc{AutoCLIP} provides a favorable choice compared to mean- and max-aggregation for many practical settings.

\section{Conclusion}
We have proposed \textsc{AutoCLIP}, a method for improving zero-shot classifiers on vision-language models. It automatically tunes per-image weights of prompt templates before aggregating them into class queries. \textsc{AutoCLIP} improves performance over standard zero-shot classifiers on the vast majority of settings, with only minimal inference-time overhead. We believe that due to its simplicity and low cost, \textsc{AutoCLIP} has the potential to be broadly applied in conjunction with vision-language models. For future work, it is exciting to explore if \textsc{AutoCLIP} also benefits other zero-shot tasks built on top of multi-modal modals such as object detection with OWL-ViT \citep{minderer2022simple} or multi-modal prompting with ImageBind \citep{Girdhar_2023_CVPR}. Moreover, exploring the potential of \textsc{AutoCLIP} for few-shot classification is an interesting direction; we present initial promising results in Section \ref{appendix:few_shot}.

\clearpage
\bibliographystyle{tmlr}
\bibliography{reference}

\clearpage
\appendix
\section{Appendix}

\begin{table*}[b]
\begin{center}
\begin{tabular}{lcccccc}
\toprule
&CLIP & CLIP & CLIP & CLIP & DataComp & CoCa\\
&RN50 & ViT-B-32 & ViT-B-16 & ViT-L-14 & ViT-L-14 & ViT-L-14\\
\midrule
              CUB200  & 47.75 (+0.1) & 53.00 (+0.4) & 57.82 (+0.3) & 64.57 (+0.3) & 85.38 (+0.4) & 73.69 (+0.1) \\
             EuroSAT  & 36.39 (-1.2) & 45.88 (-2.2) & 59.22 (+2.5) & 57.89 (+0.8) & 60.08 (-0.7) & 57.15 (-1.2) \\
             Food101  & 79.12 (+0.9) & 83.43 (+0.7) & 88.53 (+0.5) & 93.14 (+0.4) & 93.89 (+0.2) & 89.77 (+0.3) \\
         Oxford Pets  & 85.92 (+1.3) & 87.11 (+1.0) & 88.53 (+0.9) & 94.08 (+0.6) & 94.00 (+0.4) & 93.54 (+0.4) \\
            ImageNet  & 60.62 (+0.3) & 63.89 (+0.2) & 69.10 (+0.3) & 75.92 (+0.1) & 79.02 (+0.0) & 75.41 (+0.0) \\
          ImageNetV2  & 53.60 (+0.3) & 56.73 (+0.5) & 62.22 (+0.2) & 70.01 (+0.1) & 71.95 (-0.0) & 67.91 (-0.0) \\
           ImageNetR  & 28.14 (+1.3) & 49.51 (+1.6) & 58.37 (+1.7) & 73.12 (+0.6) & 78.06 (+0.7) & 73.73 (+1.1) \\
\bottomrule
\end{tabular}
        \vspace{-.25cm}
		\caption{Accuracy of \textsc{AutoCLIP} (and $\Delta$ Accuracy  to baseline zero-shot classifier in parenthesis) for $K=100$ DCLIP prompt templates across models and datasets,  averaged over 7 runs.}
		\label{Table:results_main_dclip}
\end{center}
\end{table*}

\begin{table*}[b]
\begin{center}
\begin{tabular}{lcccccc}
\toprule
&CLIP & CLIP & CLIP & CLIP & DataComp & CoCa\\
&RN50 & ViT-B-32 & ViT-B-16 & ViT-L-14 & ViT-L-14 & ViT-L-14\\
\midrule
              CUB200  & 47.00 (+0.3) & 52.36 (+0.7) & 56.99 (+1.2) & 63.94 (+0.5) & 85.52 (+1.1) & 73.99 (+0.1) \\
             EuroSAT  & 32.28 (-3.7) & 44.78 (-1.3) & 56.76 (+0.4) & 52.96 (+1.8) & 61.94 (+1.4) & 51.58 (-1.7) \\
             Food101  & 79.69 (+1.1) & 83.64 (+0.9) & 88.83 (+0.6) & 93.33 (+0.2) & 94.55 (+0.3) & 90.36 (+0.3) \\
         Oxford Pets  & 84.30 (+1.7) & 85.20 (+2.0) & 88.42 (+0.9) & 93.24 (+1.2) & 93.79 (+1.3) & 92.67 (+1.3) \\
            ImageNet  & 59.90 (+0.2) & 63.31 (+0.3) & 68.43 (+0.2) & 75.38 (+0.1) & 79.29 (+0.1) & 75.79 (+0.2) \\
          ImageNetV2  & 52.98 (+0.5) & 56.00 (+0.4) & 62.12 (+0.2) & 69.56 (-0.1) & 72.09 (+0.0) & 67.90 (-0.0) \\
           ImageNetR  & 27.11 (+0.9) & 47.74 (+0.9) & 56.28 (+1.1) & 71.30 (+0.4) & 78.26 (+0.5) & 74.51 (+0.9) \\
\bottomrule
\end{tabular}
        \vspace{-.25cm}
		\caption{Accuracy of \textsc{AutoCLIP} (and $\Delta$ Accuracy  to baseline zero-shot classifier in parenthesis) for $K=100$ CLIP prompt templates across models and datasets,  averaged over 7 runs.}
		\label{Table:results_main_clip}
\end{center}
\end{table*}

\subsection{Inference time overhead of \textsc{AutoCLIP}} \label{appendix:inference_time}
In this paragraph, we provide some measurements on inference time overhead by \textsc{AutoCLIP}.
We provide numbers for the case of a ViT-L-14 on the Oxford Pets dataset. Here, encoding an image takes $12.64$ms on a V100 (minimum over 100 images). The baseline ``averaging'' zero-shot classifiers takes additional $0.08$ms (average over 640 samples) on top to classify a sample. \textsc{AutoCLIP} takes additional $1.54$ms (average over 640 samples) for classification when running bisection for autotuning the step size. For a fixed step size, the overhead of \textsc{AutoCLIP} is 0.45ms. Thus, \textsc{AutoCLIP} with autotuning raises inference time from $12.64$ms to $14.18$ms. In contrast, TPT \citep{shu2022test} and RLCF \citep{Zhao2023TestTimeAW}, which did not report compute or memory requirements, require encoding multiple image augmentations. TPT states "We augment a single test image 63 times using random resized crops and construct a batch of 64 images, including the original one.", which means that the image encoding time (for instance the 12.64ms from above) is increased by a factor of 64x, plus additional overhead for backpropagating through the text encoder, which likely brings the inference time per sample close to $1$s (or more if multiple test-time adaptation steps are conducted). We note that for bisection, we use an independent call to \texttt{scipy.optimize.bisect} \citep{2020SciPy-NMeth} (maxiter=100, xtol=1e-2, rtol=1e-2). A batched variant of bisection could speed-up many workloads.

\subsection{Additional Experimental results} \label{appendix:additional_results}
\begin{figure*}[h]
	\begin{center}
	\includegraphics[width=\linewidth]{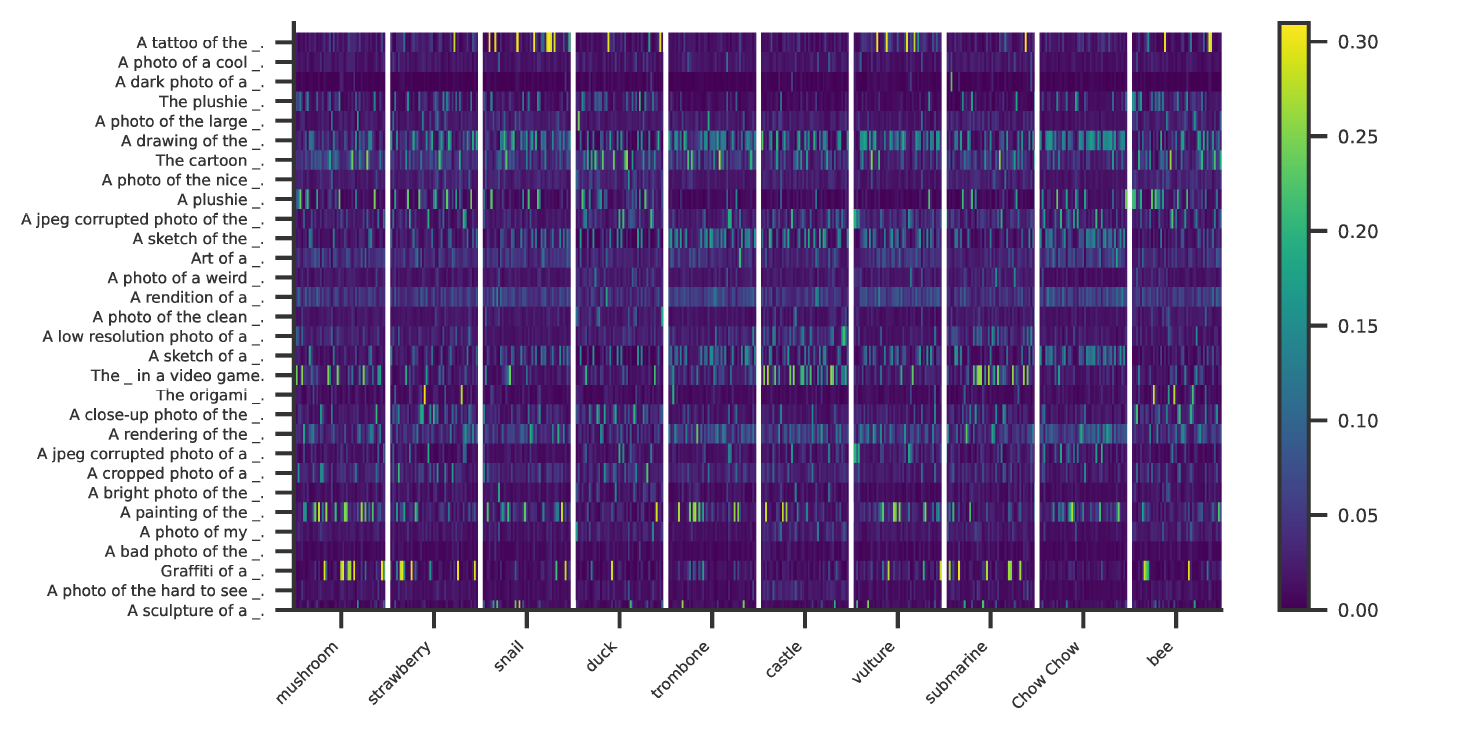} 
    \vspace*{-.5cm}
		\caption{Illustration of prompt template weights $w$ on 500 samples from the ImageNetR dataset, with blocks of 50 samples belonging to the same (unknown) class. CLIP backbone is a ViT-B-16 and 30 CLIP prompt templates are used.}
		\label{Figure:weights_imagenetr}
	\end{center}
\end{figure*}

\begin{figure*}[p]
	\begin{center}
	\includegraphics[width=\linewidth]{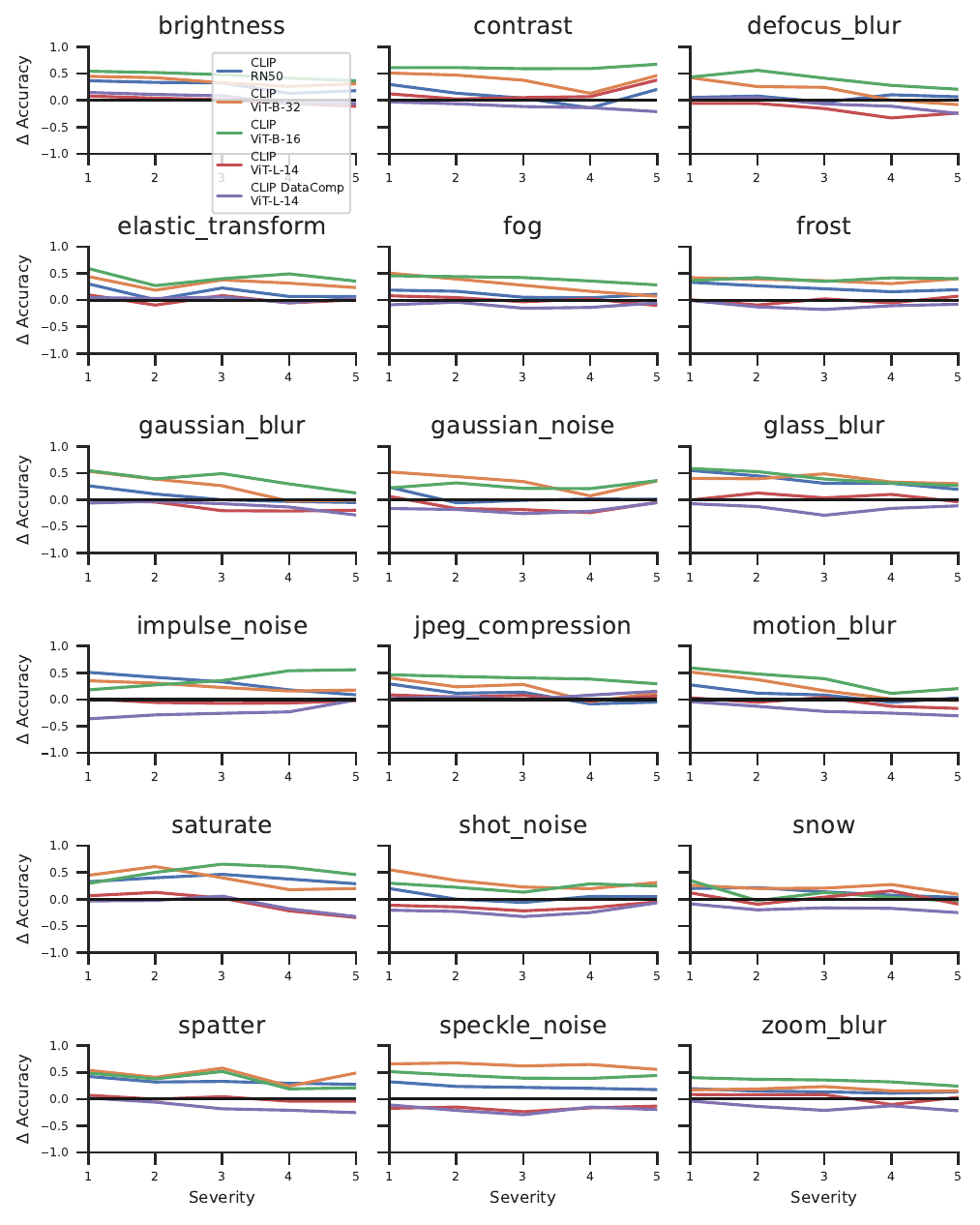} 
		\caption{ImageNetC Accuracy improvement ($\Delta$ Accuracy) of \textsc{AutoCLIP} over baseline zero-shot classifier for WaffleCLIP across models, corruptions, averaged over 7 runs.}
		\label{Figure:imagenetc_detailed}
	\end{center}
\end{figure*}

We present additional experimental results. Table \ref{Table:results_main_dclip} and Table \ref{Table:results_main_clip} show the absolute performance of \textsc{AutoCLIP} on different datasets and VLMs for DCLIP and CLIP prompt templates, respectively, similar to Table \ref{Table:results_main_waffle} in the main paper for WaffleCLIP templates. Figure \ref{Figure:weights_imagenetr} illustrates prompt weights on the ImageNetR dataset. Figure \ref{Figure:imagenetc_detailed} contains results of \textsc{AutoCLIP} in terms of $\Delta$ Accuracy on ImageNetC for every corruption seperately. 

\begin{figure*}[p]
	\begin{center}
	\includegraphics{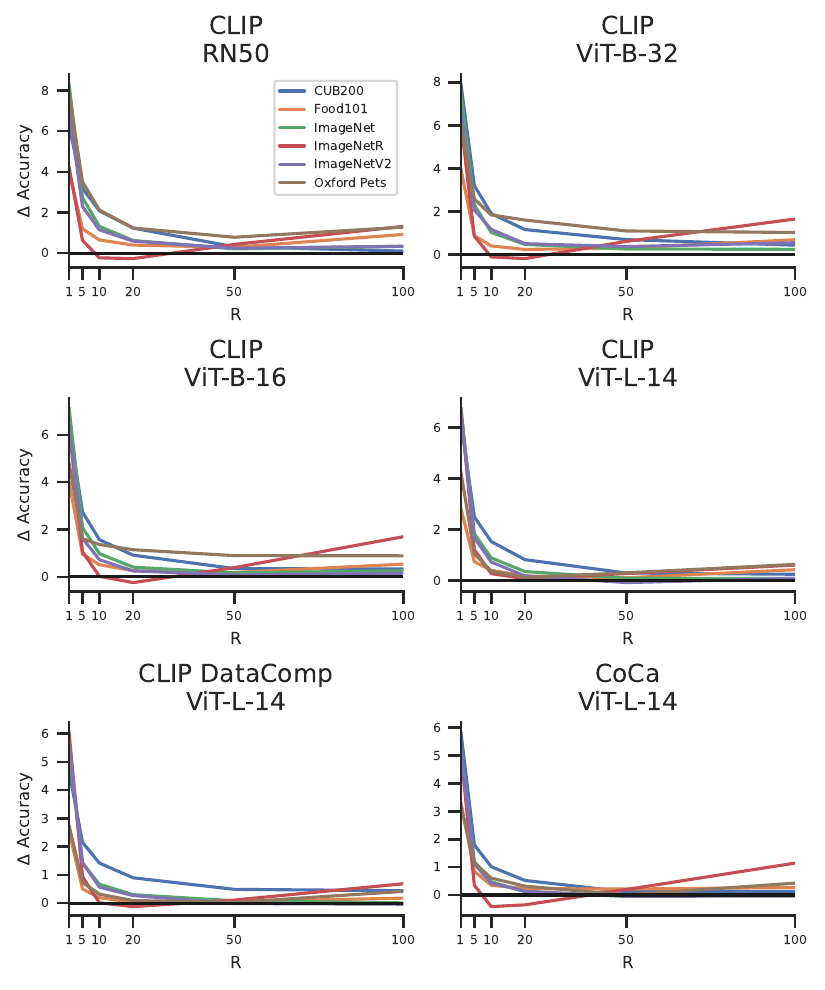} 
		\caption{Accuracy improvement ($\Delta$ Accuracy) of \textsc{AutoCLIP} with $K=100$ DCLIP prompt templates over TopR zero-shot classifier with different values of $R$ across models, averaged over datasets and 7 runs.}
		\label{Figure:results_tta_topr}
	\end{center}
\end{figure*}

In Figure \ref{Figure:results_tta_topr}, we show an additional comparison of \textsc{AutoCLIP} to a stronger baseline which is based on TopR aggregation. In this TopR aggregation, for each image $R$ prompt templates are selected whose resulting encoded class descriptors have maximum average cosine similarity to the encoded image. We note that choosing R is non-trivial in a zero-shot setting due to the lack of labelled validation data. In the figure, we compare \textsc{AutoCLIP} against this TopR-CLIP for $K=100$ DCLIP prompt template, across the same VLMs and datasets as in Figure \ref{Figure:results_main}. We provide results for different choices of $R$: overall, for the best choice of $R=20$, \textsc{AutoCLIP} is better on $86\%$ of the cases and by $0.40$ percent point accuracy on average.

\subsection{Softmax aggregation} \label{sec:softmax_aggr}
Natural baselines for \textsc{AutoCLIP} are ``mean'' aggregation ($s_j \gets 1 / K \sum_{i=1}^K  e_{ij}^{(d)} \cdot q_j$ as in Algorithm \ref{alg:zero_shot}) and ``max'' aggregation ($s_j \gets \max_i e_{ij}^{(d)} \cdot q_j$). The former corresponds to $w_i = 1 / K$ and the latter to $w_{i^*} = 1$ and $w_i =0 \quad \forall i \neq i^*$ with $i^* = \argmax_i e_{ij}^{(d)} \cdot q_j$. Naturally, it makes sense to also consider alternative ``in-between'' choices for $w$ that are (in contrast to \textsc{AutoCLIP}) not based on the gradient of some objective functions. One such choice is the ``softmax'' weighting $w = \softmax_{i=1}^K(\tau \cdot 1 / C \sum_j e_{ij}^{(d)} \cdot q_j)$, which assigns higher weight to prompt templates $t_i$ whose resulting class descriptors are more similar to the image on average. Here the temperature $\tau$ controls the sharpness of the distribution. Instead of tuning $\tau$ directly, we determine $\tau$ per example via bisection such that a specific target entropy of $w$ is obtained (analogously to \textsc{AutoCLIP} with $\beta = 0.85$).

\begin{table}[tb]
\begin{center}
\begin{tabular}{lrrrr}
\toprule
     &   AutoCLIP &    max &   mean &   softmax \\
\midrule
CUB200      &      \textbf{57.12} &  56.19 &  55.78 &     55.67 \\
EuroSAT     &      \textbf{55.93} &  55.73 &  54.53 &     55.76 \\
Food101     &      \textbf{88.85} &  88.50 &  88.00 &     88.65 \\
ImageNet    &      \textbf{68.70} &  67.06 &  68.20 &     68.40 \\
ImageNetR   &      59.13 &  56.92 &  58.15 &     \textbf{59.17} \\
ImageNetV2  &      \textbf{62.54} &  60.72 &  61.95 &     62.19 \\
Oxford Pets &      85.89 &  \textbf{87.65} &  83.99 &     82.83 \\
Average     &      \textbf{68.31} &  67.54 &  67.23 &     67.52 \\
\bottomrule
\end{tabular}
\caption{Comparison of \textsc{AutoCLIP} to different aggregation methods for a CLIP ViT-B-16 and K=100 WaffleCLIP prompt templates. Shown is accuracy, averaged over 7 runs.}
\label{Table:Aggregation}
\end{center}
\end{table}

We compare the different aggregation methods to \textsc{AutoCLIP} in Table \ref{Table:Aggregation}. ``softmax'' aggregation often (but not always) outperforms ``mean'' and ``max'' aggregation. However, AutoCLIP clearly outperforms ``softmax'' aggregation, being superior to softmax on all datasets except for ImageNetR, where both are nearly on par.

 We hypothesize that while average image-prompt template alignment contains useful information, directly determining weights based on it (such as in ``softmax' aggregation) is problematic because it does not handle entanglement between class and prompt templates (as defined in Section \ref{Section:Controlled_Setting} in our paper) well. Indeed, when evaluating ``softmax'' aggregation in the controlled setting from Figure \ref{Figure:results_toy_data}, we find that it nearly always performs between``mean'' and ``max'' aggregation (as it interpolates between the two) and thus typically worse than \textsc{AutoCLIP}.

\subsection{Motivation of Log-Sum-Exp in \textsc{AutoCLIP}} 
\label{appendix:logsumexp}
As discussed in Section \ref{Subsection:AutoCLIP}, \textsc{AutoCLIP} determines weights $w_i$ such that $\logsumexp_j(s_j) = \logsumexp_j(\sum_{i=1}^K w_i e^{(xd)}_{ij}) =\logsumexp_j(\softmax(\rho)  \cdot e^{(xd)}_{:j})$ gets increased by one step of gradient ascent in the direction of $\nabla_\rho \logsumexp_j(\softmax(\rho) \cdot e^{(xd)}_{:j})$. 
Empirically, we find in Figure \ref{Figure:results_tta_losses} that this $\logsumexp$ objective function outperforms alternative loss functions such as entropy, mean, or max. On the one hand, we attribute this to $\logsumexp$ being ``in-between'' mean and max as as a ``smooth maximum''. On the other hand, we hypothesize that $\logsumexp$ is a reasonable choice because its relationship to the energy function of a data point --- in that view \textsc{AutoCLIP} can be interpreted as minimizing the energy and maximizing the probability density $p(x)$ of $x$ under the zero-shot classifier. We elaborate this relationship in more detail in the following paragraph.

According to \citep{Grathwohl2020Your}, for a softmax-based classifier that assigns logits $f_\theta(\mathbf{x})$ to class-probablities via $p(y\vert \mathbf{x}) = \softmax(f(\mathbf{x}))$, one can slightly re-interpret the logits obtained from $f_\theta$ to
define $p(\mathbf{x}, y)$ and $p(\mathbf{x})$. For this, one re-uses the logits to define an energy-based model of the joint distribution of data point $\mathbf{x}$ and labels $y$ via $p_\theta(\mathbf{x}, y) = \dfrac{\exp(f_\theta(\mathbf{x})_{[y]})}{Z(\theta)}$, where $Z(\theta)$ is the unknown normalizing constant and $E_\theta(\mathbf{x}, y) = -f_\theta(\mathbf{x})_{[y]}$ is the energy. Moreover, $p_\theta(\mathbf{x}) = \sum_y p_\theta(\mathbf{x}, y) = \dfrac{\sum_y \exp(f_\theta(\mathbf{x})_{[y]})}{Z(\theta)}$, and accordingly the energy corresponds to $E_\theta(\mathbf{x}) = - \log \sum_y \exp(f_\theta(\mathbf{x})_{[y]})$.

In the case of \textsc{AutoCLIP}, the logits $f_\theta(\mathbf{x})$ correspond to the cosine similarities $s_j$ (with optional temperature scaling). Accordingly, $-\logsumexp_j(s_j)$ can be interpreted as the energy, and \textsc{AutoCLIP}, which maximizes $\logsumexp_j(s_j)$, can be interpreted as a method for minimizing the energy of the zero-shot classifier.

We note that this interpretation works best for classifiers that were trained with an energy-based loss term that encourages low energy on the data manifold and high energy elsewhere as discussed by \citep{Grathwohl2020Your}. However, empirically we find it to also work well on CLIP-based zero-shot classifiers.

\subsection{Step Size Selection in \textsc{AutoCLIP}} 
\label{appendix:step_size_selection}

\begin{figure}[tb]
	\begin{center}
	\includegraphics[width=.9\linewidth]{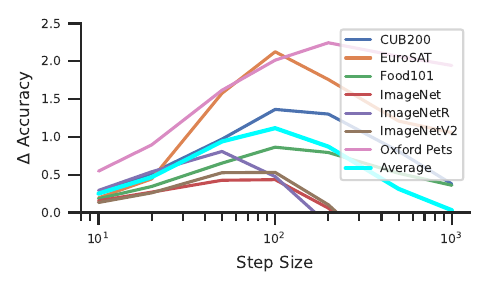} 
    \vspace{-.5cm}
		\caption{Control experiment when directly choosing a fixed step size $\alpha$ (and not a target entropy rate $\beta$). Shown is the
accuracy improvement ($\Delta$ Accuracy) of \textsc{AutoCLIP} over a baseline zero-shot classifier for a CLIP ViT-B-16, and 100 WaffleCLIP prompt templates, averaged over 7 runs.}
		\label{Figure:results_tta_step_size}
	\end{center}
\end{figure}

\textsc{AutoCLIP} controls the target entropy rate $\beta$ as free hyperparameter as an alternative parameterization to directly controlling the step size $\alpha$. The reason for this is that optimal choices of step size $\alpha$ vary more strongly across datasets than choices for the target entropy rate $\beta$. We show this in Figure \ref{Figure:results_tta_step_size} (compared to Figure \ref{Figure:results_entropy_rate}): while the reparametrization from $\alpha$ to $\beta$ does not change the peak performance of \textsc{AutoCLIP} on a dataset much, it affects the alignments of curves across datasets. We observe that the curves are more aligned when varying $\beta$ than when varying $\alpha$. Quantitatively, the mean pairwise Pearson product-moment correlation coefficient between the curves for $\beta$ in Figure \ref{Figure:results_entropy_rate} is $0.66$, while it is only $0.45$ for the curves for $\alpha$ in Figure \ref{Figure:results_tta_step_size}. Thus, in a zero-shot setting in which hyperparameters cannot be tuned on a per-dataset basis, the more aligned behavior across datasets for target entropy rate $\beta$ is preferable to the the step size $\alpha$.

\subsection{\textsc{AutoCLIP} for Few-Shot Learning} 
\label{appendix:few_shot}
While this work generally focuses on zero-shot learning settings in which no data from the task is available, it is often possible to obtain a small set of labeled data points for a task. More specifically, we consider a few-shot classification setting in which we have K labeled datapoints per class. We study \textsc{AutoCLIP} in combination with a simple CLIP-based approach for few-shot classification: in this approach, each image is considered as one exemplar of its respective class. Similar to the prompt template based approaches for zero-shot classification, this results in K encoded class descriptors, namely $e_{ij}^{(d)}$ for the i-th image of the j-th class. These encoded descriptors can be plugged into line 6 of Algorithm \ref{alg:auto_clip} as replacements for the prompt template based descriptors. This allows applying \textsc{AutoCLIP} or any other aggregation method in a few-shot setting.

\begin{figure}[tb]
	\begin{center}
	\includegraphics[width=\linewidth]{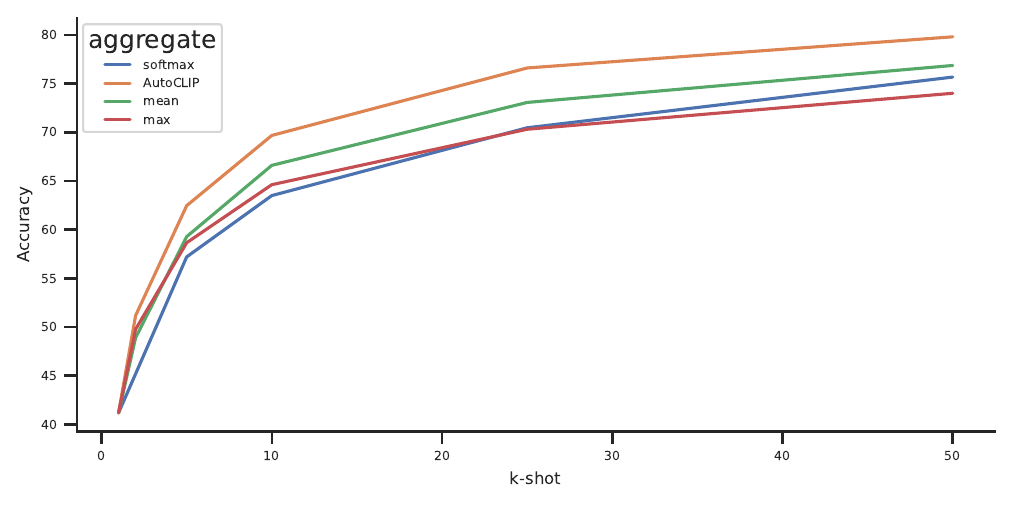} 
    \vspace{-.5cm}
		\caption{Comparison of \textsc{AutoCLIP} to other aggregation methods on few-shot CLIP-based classification on Oxfords Pets. Average over 7 runs for a CLIP ViT-B-16.}
		\label{Figure:results_few_shot}
	\end{center}
\end{figure}

Figure \ref{Figure:results_few_shot} shows the results for a CLIP ViT-B-16 on Oxford Pets for different aggregation methods. \textsc{AutoCLIP} consistently outperforms other methods by a substantial margin of 3 or more percent points accuracy. This indicates that \textsc{AutoCLIP} is a generally beneficial method for aggregating class-level information from several queries (textual or other modalities), which can be used beyond zero-shot classification. We note that while these results show promise for \textsc{AutoCLIP} in few-shot settings, future work will be needed to integrate and evaluate \textsc{AutoCLIP}  in more powerful CLIP-based few-shot learning approaches such as with the training-free Tip-Adapter \citep{Zhang2022TIP}.

\end{document}